%% file: iclr2027_conference.tex
\documentclass{article} 
\usepackage{iclr2027_conference,times}

\input{math_commands.tex}

\usepackage{hyperref}
\hypersetup{
    colorlinks=true,
    linkcolor=blue,
    citecolor=blue,
    urlcolor=magenta
}
\usepackage{url}
\usepackage{booktabs}
\usepackage{graphicx}
\usepackage{multirow}
\usepackage{subcaption}
\usepackage{natbib}
\usepackage{wrapfig}
\usepackage[table]{xcolor}

\title{RT-DETR-World: Transferring Rich LLM Semantics to Real-Time Open-Vocabulary Detection}

\author{%
\textbf{%
Yupeng Zhang\textsuperscript{1}\quad
Ziyi Zhao\textsuperscript{2}\quad
Juntao Cheng\textsuperscript{3,4}\quad
Sheng Wang\textsuperscript{1}\quad
Ningnan Guo\textsuperscript{1}}\\
\textbf{%
Ruize Han\textsuperscript{2}\quad
Liang Wan\textsuperscript{1}}\\[3pt]
\textsuperscript{1}Tianjin University
\qquad
\textsuperscript{2}Shenzhen University of Advanced Technology\\
\textsuperscript{3}Shanghai Jiaotong University
\qquad
\textsuperscript{4}Shanghai AI Laboratory\\[3pt]
{\small
\texttt{\{zhangyupeng, kevinwang630, guo\_nn, lwan\}@tju.edu.cn}}\\
{\small
\texttt{\{suat25060113, hanruize\}@suat-sz.edu.cn}}\\
{\small
\texttt{juntao-cheng@sjtu.edu.cn}}
}

\iclrfinalcopy 

\begin{document}

\maketitle

\begin{abstract}
Open-vocabulary detection (OVD) recognizes categories unseen during training through textual category queries, yet achieving strong generalization with real-time efficiency remains challenging. Beyond vocabulary scaling, zero-shot generalization may benefit from reusable visual--semantic cues learned from seen data, including attributes, actions, states, and contextual relations. Existing real-time OVD methods primarily emphasize vocabulary coverage and efficient region/query--text matching; under strict efficiency constraints, compact detectors may struggle to absorb rich instance semantics and scene context.
We propose \textbf{RT-DETR-World}, a compact DETR-style detector that transfers the rich semantics conveyed by descriptions during training while retaining lightweight query--text matching at inference. We construct \textbf{GroundingCapv2} with three levels of supervision: category names for standard OVD, object descriptions conveying instance-level semantics, and image descriptions conveying object relations and scene context. These descriptions serve only as training-time semantic supervision. To help the compact detector absorb these semantics, we propose \textbf{Dual-Path Description Alignment (DDA)}, combining a deployment-consistent MiniLM pathway with a training-only LLM teacher. MiniLM provides query--category supervision and object-description alignment, while offline teacher features supervise matched queries and global visual representations at the object and image levels, respectively. All teacher features are precomputed, and the teacher-side modules are removed after training. We further propose \textbf{Relation-Aware Negative Relaxation (RNR)}, which uses teacher-derived semantic similarities to relax related negatives while preserving exact positives. Experiments demonstrate competitive zero-shot accuracy and a favorable accuracy--efficiency trade-off. \textit{The code will be released.} 
\end{abstract}

\section{Introduction}

Open-vocabulary detection (OVD)~\citep{zareian2021open} uses natural-language queries to recognize categories unseen during training. Real-world applications such as robotics, autonomous driving, and edge vision demand strong zero-shot generalization and efficient inference. Recent real-time OVD methods, including Grounding DINO 1.5 Edge~\citep{ren2024grounding}, YOLO-World~\citep{cheng2024yolo}, YOLOE~\citep{wang2025yoloe}, and OV-DEIM~\citep{wang2026ov}, have substantially improved the speed--accuracy trade-off, highlighting the potential of compact open-vocabulary detectors.

\begin{figure*}[t]
    \centering
    \includegraphics[width=0.82\textwidth]{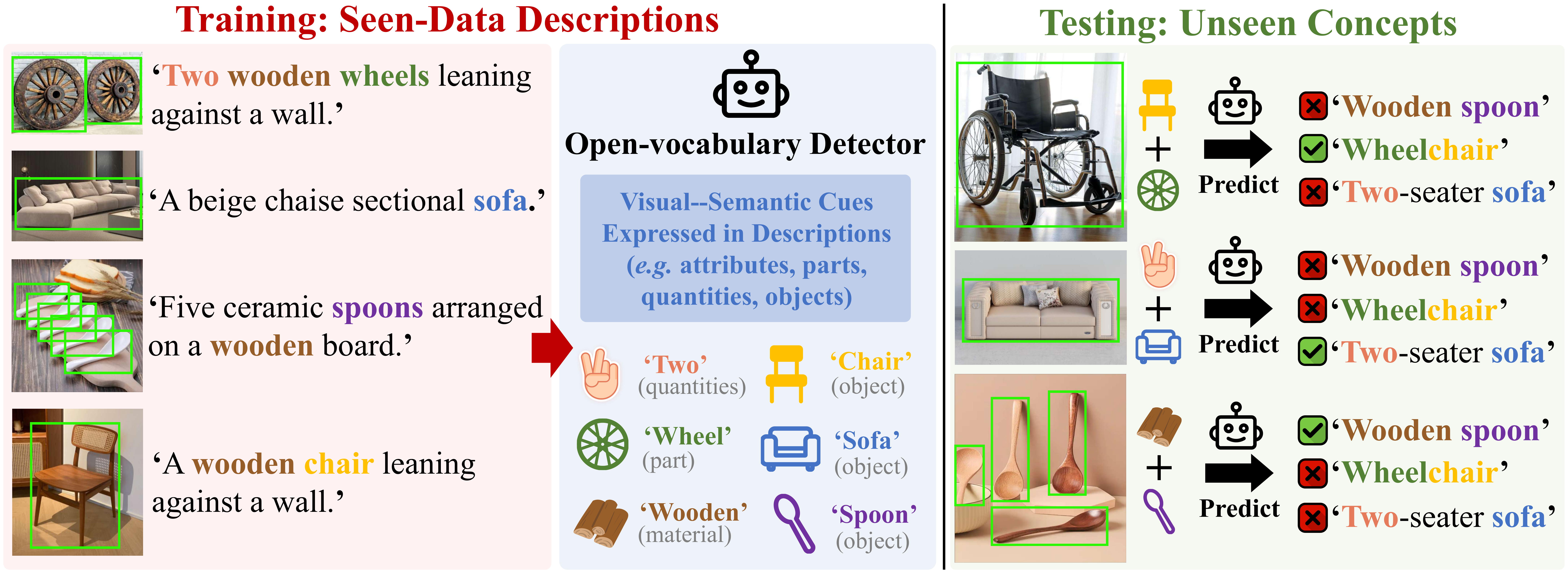}
    \vspace{-10pt}
    \caption{\textbf{Descriptions convey rich, reusable visual--semantic factors.} Fine-grained cues learned from seen data may help match visual instances to unseen category concepts at test time.}  
    \label{fig:motivation}
    \vspace{-19pt}
\end{figure*}

Beyond scaling the training vocabulary, zero-shot generalization may also benefit from reusable visual--semantic factors learned from seen data. As illustrated in Fig.~\ref{fig:motivation}, the rich semantics conveyed by descriptions reveal such fine-grained cues, helping match visual instances to unseen category concepts. For example, cues associated with \emph{chair} and \emph{wheel}, \emph{two-seat configuration} and \emph{sofa}, or \emph{wooden material} and \emph{spoon} may facilitate the recognition of \emph{wheelchair}, \emph{two-seater sofa}, and \emph{wooden spoon}, respectively. This intuition echoes the compositionality principle of compositional zero-shot learning (CZSL)~\citep{jiang2024imaginary,zhang2024disentangling}, where reusable semantic factors learned from seen compositions support the recognition of unseen combinations.

Large-scale grounded vision--language detectors further demonstrate the value of the rich semantics conveyed by descriptions, as attributes, actions, and relations provide useful cues for distinguishing and localizing object instances. GLIP~\citep{li2022grounded} and Grounding DINO~\citep{liu2024grounding} unify object detection and phrase grounding, while LLMDet~\citep{fu2025llmdet} further enriches detector representations through region- and image-level language generation. However, their heavy vision--language architectures and multi-layer cross-modal interactions hinder real-time deployment. In contrast, real-time OVD methods favor lightweight region/query--text matching. While efficient, these methods remain focused on vocabulary coverage and matching efficiency, underutilizing the rich semantics conveyed by descriptions. Moreover, under strict efficiency constraints, the limited representation capacity of compact detectors makes it difficult to fully absorb instance semantics and scene context. \textbf{Therefore, how to transfer such rich semantics into compact detectors without compromising inference efficiency remains underexplored.}

To address this challenge, we propose \textbf{RT-DETR-World}, which transfers the rich semantics conveyed by descriptions during training while retaining only lightweight vision--language matching at inference. Specifically, to obtain strong general-purpose visual representations, we adopt a frozen DINOv3~\citep{simeoni2025dinov3} as the visual encoder, followed by a lightweight multi-scale projector and a DETR decoder that produces object queries. A compact MiniLM~\citep{wang2021minilmv2} encodes category names and category-oriented textual prompts, whose embeddings are directly matched with object queries for open-vocabulary classification, avoiding the additional inference overhead of costly multi-layer cross-modal interaction. We further construct \textbf{GroundingCapv2} upon GroundingCap-1M~\citep{fu2025llmdet} with three levels of supervision: (i) \emph{category names} for standard OVD; (ii) \emph{object descriptions} conveying instance attributes, actions, poses, and states; and (iii) \emph{image descriptions} conveying object relations, scene context, and global semantics. These descriptions serve only as training-time semantic supervision, while the deployed detector retains the standard category-oriented OVD interface.

To help the compact detector absorb these multi-level semantics, we propose \textbf{Dual-Path Description Alignment (DDA)}, which combines a deployment-consistent MiniLM pathway with a training-only LLM teacher. The MiniLM pathway uses category names for standard query--category supervision and aligns object descriptions with Hungarian-matched queries to inject fine-grained instance semantics. Since lightweight alignment alone may be insufficient for a capacity-limited detector, the caption-contrastively tuned LLM text encoder from LLM2CLIP~\citep{huang2026llm2clip} provides stronger offline semantic targets: object- and image-description features supervise matched queries and global visual representations, respectively. These features are precomputed, so the LLM is absent during training and inference, and all auxiliary alignment modules are removed after training. Moreover, different descriptions may share attributes, actions, states, or context, yet standard contrastive learning treats all unmatched pairs as equally strong negatives. We therefore propose \textbf{Relation-Aware Negative Relaxation (RNR)}, which uses teacher-derived semantic similarities to reduce the repulsion of related negatives while preserving exact positives. RNR is applied to the MiniLM object-description alignment and the teacher-based object- and image-level
alignments.

Our contributions are threefold:
(1) We propose \textbf{RT-DETR-World}, a compact OVD framework that leverages rich semantic supervision from descriptions while retaining lightweight inference, and construct \textbf{GroundingCapv2} with explicit category-, object-, and image-level supervision.
(2) We introduce \textbf{DDA}, which combines a deployment-consistent MiniLM pathway with a training-only LLM teacher to help the compact detector better absorb these rich semantics without additional inference cost.
(3) We propose \textbf{RNR}, which uses teacher-derived semantic similarities to relax related negatives while preserving exact positives.
\textit{\textbf{See Appendix~\ref{sec:related_work} for related work.}}

\section{GroundingCapv2}
\label{sec:groundingcapv2}

\subsection{Tri-Level Data Formulation}

GroundingCap-1M~\citep{fu2025llmdet} aggregates localized phrases and detailed image descriptions from heterogeneous sources. Detection datasets mainly provide category labels, while grounding and image--text sources provide phrases or tags that may mix category identity, instance details, and context. We reorganize GroundingCap-1M into GroundingCapv2 with three explicit levels: category names define the standard OVD interface, while object and image descriptions convey instance- and scene-level semantics during training.
Formally, the $b$-th sample is represented as
{\footnotesize
$\mathcal{S}_b=(I_b,\mathcal{B}_b,\mathcal{T}^{\mathrm{cat}}_b,
\mathcal{T}^{\mathrm{obj}}_b,t^{\mathrm{img}}_b)$},
where $I_b$ and $\mathcal{B}_b$ denote the image and its annotated regions. The category annotations
{\footnotesize
$\mathcal{T}^{\mathrm{cat}}_b=
\{(i,c_{bi})\mid i\in\mathcal{J}^{\mathrm{cat}}_b\}$}
associate region $i$ with a concise category name $c_{bi}$, while the object annotations
{\footnotesize
$\mathcal{T}^{\mathrm{obj}}_b=
\{(i,d_{bi})\mid i\in\mathcal{J}^{\mathrm{obj}}_b\}$}
associate it with a description $d_{bi}$ conveying attributes, parts, quantities, actions, poses, and states. The image description
{\footnotesize $t^{\mathrm{img}}_b$}
captures object relations, scene context, and global semantics. Here,
{\footnotesize $\mathcal{J}^{\mathrm{cat}}_b$} and
{\footnotesize $\mathcal{J}^{\mathrm{obj}}_b$}
index regions with valid category and object-description annotations, respectively. Since they may differ, each region receives only the reliable supervision available from its source.

\setlength{\columnsep}{10pt}
\setlength{\intextsep}{4pt}
\begin{wraptable}{r}{0.37\textwidth}
    \vspace{-5pt}
    \centering
    \tiny
    \setlength{\tabcolsep}{3.0pt}
    \renewcommand{\arraystretch}{0.96}
    \caption{Statistics of GroundingCapv2 after standardization.}
    \label{tab:groundingcapv2}
    \vspace{-8pt}
    \begin{tabular}{@{}lrrrr@{}}
        \toprule
        \multirow[c]{2}{*}{Source} &
        \multirow[c]{2}{*}{Samples} &
        \multirow[c]{2}{*}{Regions} &
        \multicolumn{2}{c}{Coverage (\%)} \\
        \cmidrule(lr){4-5}
        & & & Cat. & Obj. \\
        \midrule
        COCO             & 205,815   & 1,234,235 & 100.0 & 100.0 \\
        V3Det            & 166,106   &   683,814 & 100.0 & 100.0 \\
        GQA              & 358,181   & 3,972,578 &  93.2 &  97.1 \\
        Flickr30K Ent.   &  79,035   &   531,791 &  93.9 &  99.7 \\
        LLaVA-Cap        & 306,553   & 1,628,395 &  97.6 &  99.9 \\
        \midrule
        GroundingCapv2   & 1,115,690 & 8,050,813 & 95.8 & 98.5 \\
        \bottomrule
    \end{tabular}
    \vspace{-2pt}
\end{wraptable}

Table~\ref{tab:groundingcapv2} summarizes the five sources and their region-level coverage. Cat. and Obj. denote valid category and object-description annotations, with Obj. coverage for detection-origin sources including category-name fallback. For COCO~\citep{lin2014microsoft} and V3Det~\citep{wang2023v3det}, source categories are preserved and instance-specific descriptions are generated, while unreliable descriptions fall back to category names. For GQA~\citep{hudson2019gqa}, Flickr30K Entities~\citep{plummer2015flickr30k}, and LLaVA-Cap~\citep{liu2023visual}, original grounded phrases provide object-level supervision, and category names are added only when supported by concrete visual concepts. Inherited image descriptions provide image-level supervision. Overall, GroundingCapv2 contains 1,115,690 samples and 8,050,813 regions, with category and object-description coverage of 95.8\% and 98.5\%, respectively; 1,115,656 samples also contain image descriptions.

\subsection{Confidence-Routed Data Curation}

\begin{figure*}[t]
    \centering
    \includegraphics[width=0.98\textwidth]{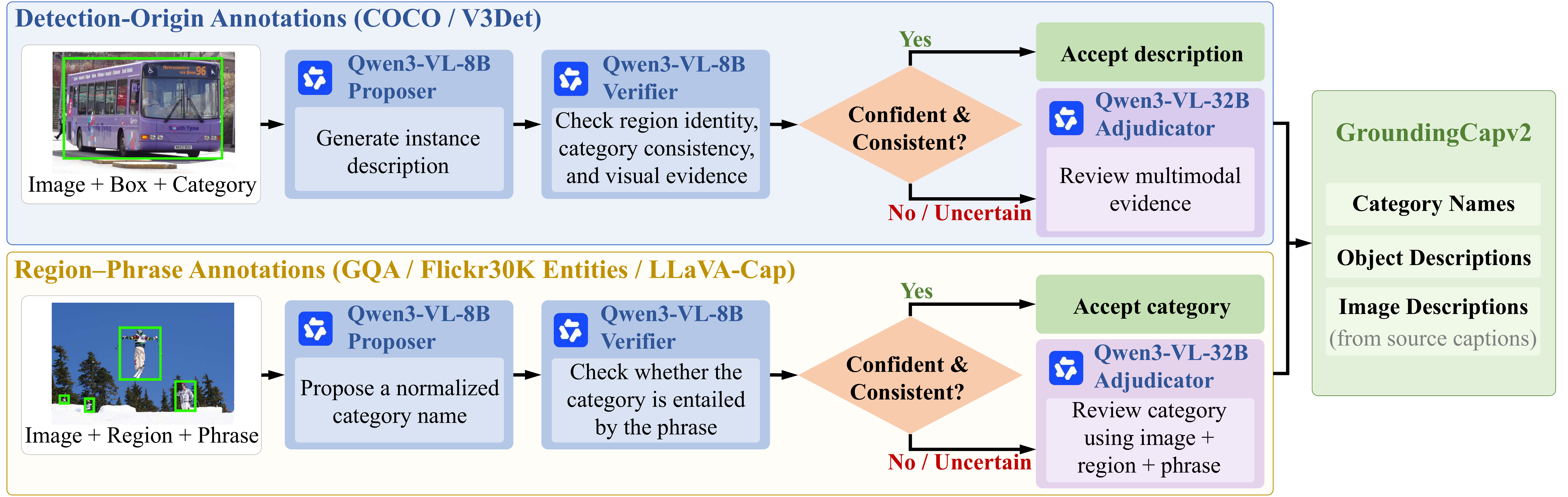}
    \vspace{-8pt}
    \caption{\textbf{Confidence-routed GroundingAgent-Opt.} Qwen3-VL-8B agents curate GroundingCap-1M category and object annotations, routing uncertain cases to Qwen3-VL-32B. Together with inherited image descriptions, curated annotations form GroundingCapv2's tri-level supervision.}
    \label{fig:groundingagent}
    \vspace{-15pt}
\end{figure*}

Generating descriptions at scale and normalizing categories may introduce incorrect visual details, category assignments, or region identities. To mitigate these errors, \textbf{GroundingAgent-Opt} employs role-specialized Qwen3-VL-8B agents for proposal and verification, while routing ambiguous cases to Qwen3-VL-32B for multimodal adjudication, as illustrated in Fig.~\ref{fig:groundingagent}. This confidence-routed workflow prioritizes annotation precision while limiting large-model use.
\textbf{(i) Detection-origin annotations.}
For COCO and V3Det, a Qwen3-VL-8B proposer generates a concise instance description from the image, annotated region, and immutable source category. An independent Qwen3-VL-8B verifier checks region identity, category consistency, and unsupported visual details. Confident and consistent descriptions are accepted, ambiguous cases are routed to Qwen3-VL-32B, and unresolved descriptions fall back to the source category.
\textbf{(ii) Region--phrase annotations.}
For GQA, Flickr30K Entities, and LLaVA-Cap, the original region phrases are retained. A text-only Qwen3-VL-8B proposer extracts and normalizes a candidate category, and an independent verifier determines whether the phrase supports a concrete visual concept. Consistent, high-confidence candidates are accepted, while ambiguous or conflicting cases are adjudicated by Qwen3-VL-32B using the image, region, and phrase. If no reliable category is identified, the category field remains empty and the original phrase continues to provide object-description supervision.

\begin{figure}[t!]
	\centering
	\includegraphics[width=0.88\linewidth]{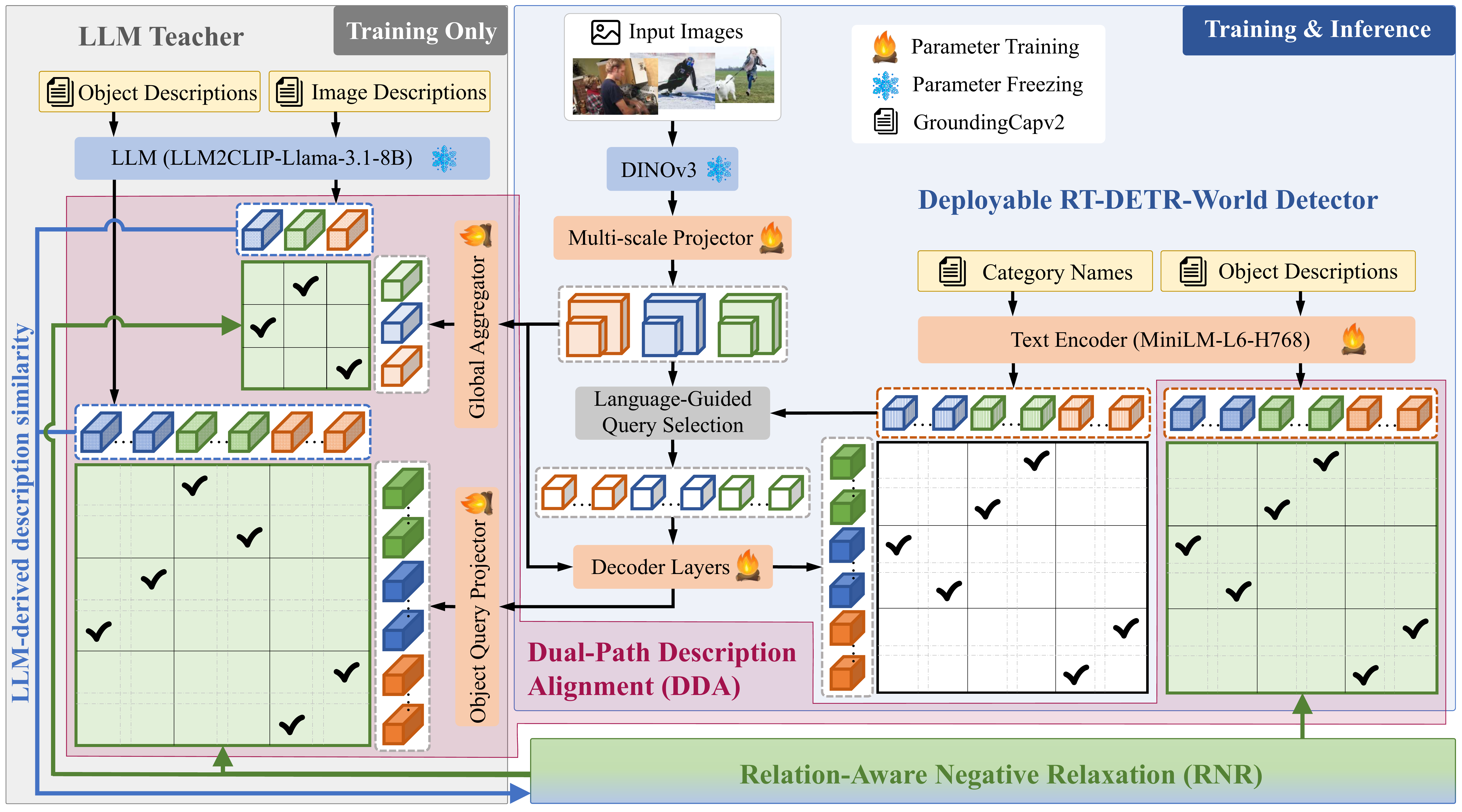}  \vspace{-10pt}
    \caption{\textbf{Overview of RT-DETR-World.} A frozen DINOv3 visual encoder, lightweight multi-scale projector, DETR decoder, and compact MiniLM form the deployable detector. DDA combines a deployment-consistent MiniLM pathway with training-only object- and image-level LLM supervision. RNR uses LLM-derived description similarities to relax semantically related negatives across the three description-alignment objectives. All teacher-side components are removed at inference.}  
	\label{fig:overview}
	\vspace{-15pt}
\end{figure}

\paragraph{Quality control.}
COCO and V3Det source categories remain fixed. GroundingAgent-Opt accepts generated descriptions for 85.5\% and 87.6\% of their regions, respectively, with the remainder using category-name fallback. All original phrases from GQA, Flickr30K Entities, and LLaVA-Cap are retained. We rebuild the category- and object-description-to-region mappings, verify region indices, token spans, and reconstructed text, and retain confidence and routing metadata solely for auditing.


\section{RT-DETR-World}
\label{sec:method}

\subsection{Overall Architecture}

As shown in Fig.~\ref{fig:overview}, RT-DETR-World comprises a compact deployable detector and training-only alignment branches supervised by precomputed features from an offline LLM teacher. Given an image, a frozen DINOv3~\citep{simeoni2025dinov3} extracts visual features, which are adapted into multi-scale detection features by a lightweight projector. MiniLM~\citep{wang2021minilmv2} encodes category names and category-oriented textual prompts. Vision--text similarities guide Language-Guided Query Selection and open-vocabulary classification, while the DETR decoder refines the selected queries and regresses bounding boxes.
During training, \textbf{Dual-Path Description Alignment (DDA)} complements standard query--category supervision with the rich semantics conveyed by object and image descriptions. Its deployment-consistent MiniLM pathway aligns object descriptions with Hungarian-matched queries, while the teacher pathway aligns matched queries and global visual representations with precomputed LLM2CLIP embeddings of object and image descriptions, respectively. \textbf{Relation-Aware Negative Relaxation (RNR)} uses teacher-derived semantic similarities to reduce the repulsion of related negatives across the three alignment objectives while preserving exact positives.
The LLM itself is absent during detector training and inference, and all auxiliary alignment modules are removed after training. The deployed detector retains only the lightweight, category-oriented MiniLM interface.



\subsection{Dual-Path Description Alignment (DDA)}
\label{sec:dda}

The rich semantics conveyed by object and image descriptions provide dense supervision, yet a compact real-time detector may struggle to absorb them through a single lightweight query--text pathway. We therefore propose \textbf{Dual-Path Description Alignment (DDA)}, which combines a deployment-consistent MiniLM pathway with a training-only teacher pathway based on precomputed LLM features. The teacher pathway provides expressive semantic targets, while the MiniLM pathway transfers fine-grained instance semantics into the representations retained at inference. All auxiliary alignment modules are removed after training, adding no inference overhead.

\paragraph{Offline LLM semantic targets.}
Following the notation in Sec.~\ref{sec:groundingcapv2}, let $b$ index a sample and $i$ an annotated region in $\mathcal{B}_b$. DDA encodes the semantics conveyed by descriptions using the caption-contrastively tuned LLM text encoder from LLM2CLIP~\citep{huang2026llm2clip}. For each $i\in\mathcal{J}_b^{\mathrm{obj}}$, the object description is encoded as $\mathbf{e}_{bi}^{\mathrm{obj}}=E_{\mathrm{L}}(d_{bi})$, while each available image description yields $\mathbf{e}_{b}^{\mathrm{img}}=E_{\mathrm{L}}(t_b^{\mathrm{img}})$. Here, $E_{\mathrm{L}}$ denotes the frozen LLM text encoder, and $\mathbf{e}_{bi}^{\mathrm{obj}},\mathbf{e}_{b}^{\mathrm{img}}\in\mathbb{R}^{d_{\mathrm{L}}}$ are the corresponding embeddings in the $d_{\mathrm{L}}$-dimensional teacher space. They serve as semantic alignment targets and relation references for RNR. All teacher embeddings are precomputed; thus, the LLM is never loaded during detector training or inference.

\paragraph{Deployment-consistent MiniLM pathway.}
Let $\mathbf{q}_{b,\pi_b(i)}\in\mathbb{R}^{d_{\mathrm{q}}}$ denote the final-layer object query assigned by Hungarian matching to region $i$ in sample $b$, where $\pi_b$ is the assignment function and $d_{\mathrm{q}}$ is the query dimension. For each $i\in\mathcal{J}_b^{\mathrm{cat}}$, MiniLM encodes the category name $c_{bi}$ for the standard query--category classification objective. For each $i\in\mathcal{J}_b^{\mathrm{obj}}$, we pool the valid non-special-token features of $d_{bi}$ and project the text and query representations into a shared alignment space:
\begingroup
\footnotesize
\begin{equation}
\mathbf{m}_{bi}
=
P_{\mathrm{m}}^{t}\!\left(
\operatorname{Pool}\!\left(E_{\mathrm{m}}(d_{bi})\right)
\right),
\qquad
\mathbf{v}_{bi}^{\mathrm{m}}
=
P_{\mathrm{m}}^{v}\!\left(
\mathbf{q}_{b,\pi_b(i)}
\right),
\label{eq:minilm-alignment}
\end{equation}
\endgroup
where $E_{\mathrm{m}}(d_{bi})\in\mathbb{R}^{n_{bi}\times d_{\mathrm{m}}}$ denotes the MiniLM token-feature sequence, with sequence length $n_{bi}$ and hidden dimension $d_{\mathrm{m}}$. $\operatorname{Pool}$ averages the valid non-special-token features. The learnable projections $P_{\mathrm{m}}^{t}:\mathbb{R}^{d_{\mathrm{m}}}\rightarrow\mathbb{R}^{d_{\mathrm{a}}}$ and $P_{\mathrm{m}}^{v}:\mathbb{R}^{d_{\mathrm{q}}}\rightarrow\mathbb{R}^{d_{\mathrm{a}}}$ map the text and query features into a shared $d_{\mathrm{a}}$-dimensional space. This alignment enriches the MiniLM and object-query representations retained at inference with fine-grained instance semantics. Object descriptions and both projection heads are used only during training.

\paragraph{Training-only teacher alignment.}
The teacher pathway aligns object- and image-level visual representations with the precomputed semantic targets. For each $i\in\mathcal{J}_b^{\mathrm{obj}}$, the Hungarian-matched query is projected into the teacher space. At the image level, valid spatial features from each projected feature level are pooled, concatenated, and mapped into the same space:
\begingroup
\footnotesize
\begin{equation}
\mathbf{v}_{bi}^{\mathrm{tch}}
=
P_{\mathrm{o}}\!\left(
\mathbf{q}_{b,\pi_b(i)}
\right),
\qquad
\mathbf{g}_{b}
=
P_{\mathrm{g}}\!\left(
\mathop{\Vert}_{\ell=1}^{L}
\operatorname{MAP}\!\left(\mathbf{F}_{b}^{\ell}\right)
\right),
\label{eq:teacher-alignment}
\end{equation}
\endgroup
where $P_{\mathrm{o}}:\mathbb{R}^{d_{\mathrm{q}}}\rightarrow\mathbb{R}^{d_{\mathrm{L}}}$ maps the matched query into the teacher space, yielding $\mathbf{v}_{bi}^{\mathrm{tch}}$. $\mathbf{F}_{b}^{\ell}\in\mathbb{R}^{H_{\ell}\times W_{\ell}\times d_{\ell}}$ denotes the projected visual feature at level $\ell$, and $L$ is the number of feature levels. $\operatorname{MAP}$ performs mask-aware average pooling over valid spatial positions, while $\Vert$ denotes concatenation across feature levels. Finally, $P_{\mathrm{g}}:\mathbb{R}^{\sum_{\ell=1}^{L}d_{\ell}}\rightarrow\mathbb{R}^{d_{\mathrm{L}}}$ maps the concatenated feature into the teacher space, yielding the global visual representation $\mathbf{g}_{b}$. We align $\mathbf{v}_{bi}^{\mathrm{tch}}$ with $\mathbf{e}_{bi}^{\mathrm{obj}}$ and $\mathbf{g}_{b}$ with $\mathbf{e}_{b}^{\mathrm{img}}$ to transfer instance semantics and global scene semantics, respectively. Both teacher-alignment projection heads are discarded after training.

\subsection{Relation-Aware Negative Relaxation (RNR)}
\label{sec:rnr}

The semantics conveyed by descriptions are not mutually exclusive: unmatched samples may share attributes, actions, states, or context, yet standard contrastive learning weights all unmatched cross-modal pairs equally, potentially over-separating related representations. We therefore propose \textbf{Relation-Aware Negative Relaxation (RNR)}, which uses precomputed LLM2CLIP embeddings to estimate semantic relations and selectively reduces the repulsion of related negatives while preserving exact positives. 
Owing to its stronger modeling of complex descriptions, the semantic similarities derived from the caption-contrastively tuned LLM teacher embeddings provide more reliable estimates of paraphrases, implicit relations, and partial semantic overlap than surface-form similarity or similarities derived from the evolving MiniLM representations.
Freezing and precomputing its embeddings further ensures consistent relation estimates without online LLM inference.

For each DDA alignment objective, we collect $N$ triplets
$\{(\mathbf{v}_n,\mathbf{t}_n,\mathbf{e}_n)\}_{n=1}^{N}$,
where $\mathbf{v}_n$ and $\mathbf{t}_n$ are paired visual and text representations, and $\mathbf{e}_n$ is the corresponding frozen teacher embedding. The bidirectional logits are
\begingroup
\footnotesize
\begin{equation}
z_{nm}^{\mathrm{v2t}}
=
\frac{\operatorname{sim}(\mathbf{v}_n,\mathbf{t}_m)}{\tau},
\qquad
z_{nm}^{\mathrm{t2v}}
=
\frac{\operatorname{sim}(\mathbf{t}_n,\mathbf{v}_m)}{\tau},
\label{eq:rnr-logits}
\end{equation}
\endgroup
where $\operatorname{sim}(\cdot,\cdot)$ denotes cosine similarity, $\tau>0$ is the temperature, and $n,m\in\{1,\ldots,N\}$ index the paired representations.

The teacher-derived semantic relation and its corresponding pair weight are defined as
\begingroup
\footnotesize
\begin{equation}
r_{nm}
=
\left[\operatorname{sim}(\mathbf{e}_n,\mathbf{e}_m)\right]_{+},
\qquad
w_{nm}
=
\begin{cases}
1, & m\in\mathcal{P}_n,\\
1-\alpha r_{nm}, & m\notin\mathcal{P}_n,
\end{cases}
\label{eq:rnr-weight}
\end{equation}
\endgroup
where $[x]_{+}=\max(x,0)$ and $r_{nm}\in[0,1]$. $\mathcal{P}_n$ denotes the exact positive correspondences of anchor $n$, typically the singleton $\{n\}$, and $\alpha\in[0,1]$ controls the maximum relaxation. Exact positives retain unit weight, while related negatives receive weights in $[1-\alpha,1]$.

For either direction $s\in\{\mathrm{v2t},\mathrm{t2v}\}$, the per-anchor loss is
$\ell_n^{s}=-\log\!\left[
\sum_{m\in\mathcal{P}_n}\exp(z_{nm}^{s})
\big/
\sum_{m=1}^{N}w_{nm}\exp(z_{nm}^{s})
\right]$.
The complete bidirectional objective is
$\mathcal{C}_{\mathrm{RNR}}(\mathbf{V},\mathbf{T};\mathbf{E})
=\tfrac{1}{2N}\sum_{n=1}^{N}
(\ell_n^{\mathrm{v2t}}+\ell_n^{\mathrm{t2v}})$,
where $\mathbf{V}$, $\mathbf{T}$, and $\mathbf{E}$ stack the visual, text, and teacher representations, respectively.

RNR leaves exact-positive terms unchanged and discounts only the denominator contributions of semantically related negatives. Thus, an unmatched pair is relaxed but never treated as a positive. The same teacher-derived relation weights are used in both directions, and $\alpha=0$ recovers standard bidirectional contrastive learning.

We apply RNR to the three DDA alignment objectives:
\begingroup
\footnotesize
\begin{equation}
\mathcal{L}_{\mathrm{M}}^{\mathrm{obj}}
=
\mathcal{C}_{\mathrm{RNR}}\!\left(
\mathbf{V}^{\mathrm{m}},\mathbf{M};\mathbf{E}^{\mathrm{obj}}
\right),
\quad
\mathcal{L}_{\mathrm{L}}^{\mathrm{obj}}
=
\mathcal{C}_{\mathrm{RNR}}\!\left(
\mathbf{V}^{\mathrm{tch}},\mathbf{E}^{\mathrm{obj}};\mathbf{E}^{\mathrm{obj}}
\right),
\quad
\mathcal{L}_{\mathrm{L}}^{\mathrm{img}}
=
\mathcal{C}_{\mathrm{RNR}}\!\left(
\mathbf{G},\mathbf{E}^{\mathrm{img}};\mathbf{E}^{\mathrm{img}}
\right),
\label{eq:dda-rnr-losses}
\end{equation}
\endgroup
where $\mathbf{V}^{\mathrm{m}}$, $\mathbf{M}$,
$\mathbf{V}^{\mathrm{tch}}$, and $\mathbf{E}^{\mathrm{obj}}$
stack the valid object-level representations over all $(b,i)$ with
$i\in\mathcal{J}_b^{\mathrm{obj}}$, while $\mathbf{G}$ and
$\mathbf{E}^{\mathrm{img}}$ stack image-level representations for samples
with valid image descriptions. In the MiniLM objective,
$\mathbf{E}^{\mathrm{obj}}$ determines only the relation weights; in the
teacher objectives, the LLM embeddings serve as both alignment targets and
relation references.

\subsection{Training Objective and Implementation Details}
\label{sec:training-objectives}

\paragraph{Training objective.}
The base detection objective $\mathcal{L}_{\mathrm{det}}$ includes Hungarian-matched query--category classification and box regression losses. Combining it with three DDA objectives gives
\begingroup
\footnotesize
\begin{equation}
\mathcal{L}
=
\mathcal{L}_{\mathrm{det}}
+
\lambda_{\mathrm{M}}\mathcal{L}_{\mathrm{M}}^{\mathrm{obj}}
+
\lambda_{\mathrm{o}}\mathcal{L}_{\mathrm{L}}^{\mathrm{obj}}
+
\lambda_{\mathrm{g}}\mathcal{L}_{\mathrm{L}}^{\mathrm{img}},
\label{eq:total-objective}
\end{equation}
\endgroup
where $\lambda_{\mathrm{M}}$, $\lambda_{\mathrm{o}}$, and $\lambda_{\mathrm{g}}$ weight the MiniLM object, LLM object, and LLM image objectives, respectively. RNR introduces no additional loss term and only reweights negative contributions within these description-alignment objectives. At inference, offline teacher targets and training-only alignment modules are discarded, while the deployable MiniLM query--text pathway remains unchanged.


\paragraph{Implementation details.}
RT-DETR-World-S, -S+, and -B use frozen DINOv3 ViT-S/16, ViT-S+/16, and ViT-B/16 backbones~\citep{simeoni2025dinov3}, respectively. Each variant employs a two-layer multi-scale projector with 256-dimensional features and a three-layer DETR decoder. We use 13 groups of 300 object queries during training and retain one 300-query group at inference. MiniLMv2-L6-H768~\citep{wang2021minilmv2} has a maximum input length of 256 tokens, while the LLM2CLIP teacher provides precomputed 1,152-dimensional object- and image-description embeddings. We set $(\lambda_{\mathrm{M}},\lambda_{\mathrm{o}},\lambda_{\mathrm{g}})=(0.25,0.1,0.1)$ in Eq.~(\ref{eq:total-objective}), $\tau=0.07$ in Eq.~(\ref{eq:rnr-logits}), and $\alpha=0.25$ in Eq.~(\ref{eq:rnr-weight}). The three description-alignment losses are linearly warmed up over the first 2,000 iterations.

All variants are trained for 150k iterations on four NVIDIA RTX 4090 GPUs with 12 images per GPU, yielding a batch size of 48. During training, the shorter image side is randomly resized within $[480,800]$ pixels, with the longer side capped at 1,333 pixels. At evaluation, the shorter side is resized to 800 pixels with the same cap. We use AdamW with a learning rate of $3\times10^{-4}$ and weight decay of $10^{-4}$; MiniLMv2 uses a learning-rate multiplier of $0.1$. The learning rate is warmed up for 1k iterations and multiplied by $0.1$ at 120k and 140k iterations.

\begin{table*}[t]
\centering
\caption{Zero-shot detection on LVIS minival, reported as standard AP / Fixed AP.}
\label{tab:lvis}
\vspace{-7pt}
\setlength{\tabcolsep}{2.0pt}
\renewcommand{\arraystretch}{1.08}
\resizebox{\textwidth}{!}{
\begin{tabular}{lllccccccccc}
\toprule
\multirow[c]{2}{*}{Method}
& \multirow[c]{2}{*}{Backbone}
& \multirow[c]{2}{*}{Text Encoder}
& \multirow[c]{2}{*}{Training Data}
& \multirow[c]{2}{*}{FPS}
& \multirow[c]{2}{*}{$\mathrm{AP}/\mathrm{AP}^{\mathrm{Fixed}}$}
& \multirow[c]{2}{*}{$\mathrm{AP}_{r}/\mathrm{AP}_{r}^{\mathrm{Fixed}}$}
& \multirow[c]{2}{*}{$\mathrm{AP}_{c}/\mathrm{AP}_{c}^{\mathrm{Fixed}}$}
& \multirow[c]{2}{*}{$\mathrm{AP}_{f}/\mathrm{AP}_{f}^{\mathrm{Fixed}}$}
& \multicolumn{3}{c}{Parameters (M)} \\
\cmidrule(lr){10-12}
&&&&&&&&& Visual/Det. & Text & Total \\
\midrule

GLIP-T   
& Swin-T & BERT-base & OG & --
& -- / 24.9 & -- / 17.7 & -- / 19.5 & -- / 31.0
& $\approx$122 & $\approx$110 & 232 \\

GLIP-T
& Swin-T & BERT-base & OG, Cap4M & --
& -- / 26.0 & -- / 20.8 & -- / 21.4 & -- / 31.0
& $\approx$122 & $\approx$110 & 232 \\

GLIPv2-T   
& Swin-T & BERT-base & OG, Cap4M & --
& -- / 29.0 & -- / -- & -- / -- & -- / --
& $\approx$122 & $\approx$110 & 232 \\

Grounding DINO-T   
& Swin-T & BERT-base & OG & --
& -- / 25.6 & -- / 14.4 & -- / 19.6 & -- / 32.2
& $\approx$62 & $\approx$110 & 172 \\

Grounding DINO-T
& Swin-T & BERT-base & OG, Cap4M & --
& -- / 27.4 & -- / 18.1 & -- / 23.3 & -- / 32.7
& $\approx$62 & $\approx$110 & 172 \\

MM-Grounding-DINO-T   
& Swin-T & BERT-base & O365, GoldG, GRIT, V3Det & --
& -- / 41.4 & -- / 34.2 & -- / 37.4 & -- / 46.2
& 64.1 & 108.9 & 173.0 \\

LLMDet-T   
& Swin-T & BERT-base & GroundingCap-1M & --
& -- / 44.7 & -- / 37.3 & -- / 39.5 & -- / 50.7
& 64.1 & 108.9 & 173.0 \\

\midrule

Grounding DINO 1.5 Edge   
& EfficientViT-L1 & -- & G-20M & --
& -- / 33.5 & -- / 28.0 & -- / 34.3 & -- / 33.9
& -- & -- & -- \\

YOLO-Worldv2.1-S   
& YOLOv8-S & CLIP-B/32 & OG & 18$^{*}$
& -- / 25.5 & -- / 19.1 & -- / 22.6 & -- / 29.3
& 13 & 63.4$^{\dagger}$ & 76.4 \\

YOLO-Worldv2.1-M
& YOLOv8-M & CLIP-B/32 & OG & 17$^{*}$
& -- / 32.7 & -- / 24.4 & -- / 30.2 & -- / 36.4
& 29 & 63.4$^{\dagger}$ & 92.4 \\

YOLO-Worldv2.1-L
& YOLOv8-L & CLIP-B/32 & OG & 16$^{*}$
& -- / 35.5 & -- / 24.4 & -- / 34.0 & -- / 38.8
& 48 & 63.4$^{\dagger}$ & 111.4 \\

YOLOEv8-S   
& YOLOv8-S & MobileCLIP-B(LT) & OG & 18$^{*}$
& 25.7 / 27.9 & 19.0 / 22.3 & 25.9 / 27.8 & 26.7 / 29.0
& 12 & 63.4$^{\dagger}$ & 75.4 \\

YOLOEv8-M
& YOLOv8-M & MobileCLIP-B(LT) & OG & 17$^{*}$
& 29.9 / 32.6 & 23.6 / 26.9 & 29.2 / 31.9 & 31.7 / 34.4
& 27 & 63.4$^{\dagger}$ & 90.4 \\

YOLOEv8-L
& YOLOv8-L & MobileCLIP-B(LT) & OG & 17$^{*}$
& 33.3 / 35.9 & 30.8 / 33.2 & 32.2 / 34.8 & 34.6 / 37.3
& 45 & 63.4$^{\dagger}$ & 108.4 \\

YOLOEv11-S
& YOLOv11-S & MobileCLIP-B(LT) & OG & 18$^{*}$
& 25.2 / 27.5 & 19.3 / 21.4 & 24.4 / 26.8 & 26.7 / 29.3
& 10 & 63.4$^{\dagger}$ & 73.4 \\

YOLOEv11-M
& YOLOv11-M & MobileCLIP-B(LT) & OG & 18$^{*}$
& 30.5 / 33.0 & 22.4 / 26.9 & 30.4 / 32.5 & 32.1 / 34.5
& 21 & 63.4$^{\dagger}$ & 84.4 \\

YOLOEv11-L
& YOLOv11-L & MobileCLIP-B(LT) & OG & 17$^{*}$
& 32.4 / 35.2 & 25.6 / 29.1 & 31.9 / 35.0 & 34.1 / 36.5
& 26 & 63.4$^{\dagger}$ & 89.4 \\

OV-DEIM-S  
& ViT-T & MobileCLIP-B(LT) & OG & 161
& 27.7 / 29.6 & 23.6 / 25.2 & 28.1 / 30.2 & 28.0 / 30.0
& 11 & 63.4$^{\dagger}$ & 74.4 \\

OV-DEIM-M
& ViT-T+ & MobileCLIP-B(LT) & OG & 109
& 30.6 / 32.6 & 25.3 / 26.9 & 30.2 / 31.5 & 31.9 / 34.1
& 20 & 63.4$^{\dagger}$ & 83.4 \\

OV-DEIM-L
& ViT-S & MobileCLIP-B(LT) & OG & 91
& 33.7 / 35.9 & 34.3 / 36.8 & 33.4 / 35.5 & 34.0 / 36.0
& 36 & 63.4$^{\dagger}$ & 99.4 \\

\midrule

\rowcolor{blue!8}
\textbf{RT-DETR-World-S (Ours)}
& DINOv3-S/16 & MiniLMv2-L6-H768 & GroundingCapv2 & 90
& 31.4 / 35.2
& 28.2 / 32.0
& 30.4 / 33.8
& 32.9 / 36.9
& 36.5 & 66.6 & 103.1 \\

\rowcolor{blue!8}
\textbf{RT-DETR-World-S+ (Ours)}
& DINOv3-S+/16 & MiniLMv2-L6-H768 & GroundingCapv2 & 71
& 32.6 / 36.6
& 26.6 / 31.1
& 31.9 / 35.8
& 34.3 / 38.2
& 43.6 & 66.6 & 110.2 \\

\rowcolor{blue!8}
\textbf{RT-DETR-World-B (Ours)}
& DINOv3-B/16 & MiniLMv2-L6-H768 & GroundingCapv2 & 41
& 36.6 / 40.0
& 33.7 / 37.1
& 36.4 / 39.6
& 37.4 / 40.9
& 120.6 & 66.6 & 187.2 \\

\bottomrule
\end{tabular}
}
\begin{minipage}{0.99\textwidth}
\tiny
\textbf{Notes:}
``--'' denotes an unreported result.
OG denotes Objects365 and GoldG, O365 denotes Objects365, and G-20M denotes Grounding-20M.
Visual/Det. denotes deployable detector parameters excluding the text encoder, while Total includes both.
For GLIP, GLIPv2, and the original Grounding DINO, the Visual/Det.--text decomposition is estimated from reported model sizes and encoder specifications.
The text encoder and complete parameter breakdown of Grounding DINO 1.5 Edge are not publicly reported.
$^{\dagger}$ The text encoder is used to construct or cache vocabulary embeddings and is excluded from the detector parameters reported by the original method; we report it separately and include it in Total.
Training-only language modules, including the LLM branch of LLMDet and our LLM2CLIP teacher, are excluded.
FPS follows the OV-DEIM protocol with category embeddings cached before timing.
$^{*}$ FPS includes NMS post-processing overhead.
\end{minipage}
\vspace{-15pt}
\end{table*}

\section{Experimental Results}

\subsection{Setup}

\paragraph{Evaluation protocol and metrics.}
All RT-DETR-World variants are trained on GroundingCapv2 described in Sec.~\ref{sec:groundingcapv2} and evaluated without target-specific fine-tuning. For LVIS v1.0 minival~\citep{gupta2019lvis}, we use the official names of all 1,203 categories as category queries without training on LVIS annotations. We report standard AP and Fixed AP~\citep{dave2021evaluating}, together with their rare-, common-, and frequent-category counterparts. These groups contain 337, 461, and 405 categories, respectively. Standard AP uses the default limit of 300 detections per image, whereas Fixed AP replaces this image-level limit with a fixed per-category detection budget. 
Following prior OVD implementations, we retain the top 1,000 predictions over all query--category pairs for Fixed AP evaluation.
Unless otherwise specified, the remaining benchmarks report standard AP.
For efficiency evaluation, we follow OV-DEIM~\citep{wang2026ov}. 
FPS is measured on a T4 GPU using TensorRT, with category embeddings cached and NMS included where applicable.

\paragraph{Comparison methods.}
We compare RT-DETR-World with two groups of open-vocabulary detectors. The first comprises general-purpose OVD and grounding models, including GLIP~\citep{li2022grounded}, GLIPv2~\citep{zhang2022glipv2}, Grounding DINO~\citep{liu2024grounding}, MM-Grounding-DINO~\citep{zhao2024open}, and LLMDet~\citep{fu2025llmdet}. The second comprises real-time OVD models, including Grounding DINO 1.5 Edge~\citep{ren2024grounding}, YOLO-World~\citep{cheng2024yolo}, YOLOE~\citep{wang2025yoloe}, and OV-DEIM~\citep{wang2026ov}. Unless otherwise stated, baseline results are taken from the corresponding papers or official repositories.

\subsection{Transfer and Robustness Evaluation}
\label{sec:transfer-evaluation}

We evaluate zero-shot transfer on LVIS minival~\citep{gupta2019lvis} and ODinW13/35~\citep{li2022elevater}, and distribution-shift robustness on COCO-O~\citep{mao2023coco}, using COCO~\citep{lin2014microsoft} as the in-distribution reference. Together, these benchmarks assess large-vocabulary detection, cross-domain transfer, and robustness to natural distribution shifts. To prevent evaluation leakage, we exclude from GroundingCapv2 any images overlapping the target evaluation splits. Since GroundingCapv2 contains COCO training images and COCO-O shares the COCO category space, COCO-O is treated as a distribution-shift benchmark rather than unseen-category evaluation. At inference, the LLM teacher and all training-only alignment modules are removed, leaving only the deployable RT-DETR-World detector.

\paragraph{Large-vocabulary detection on LVIS minival.}
LVIS minival contains 1,203 categories with a long-tailed distribution over rare, common, and frequent classes. As shown in Table~\ref{tab:lvis}, RT-DETR-World-S, -S+, and -B achieve standard/Fixed AP scores of 31.4/35.2, 32.6/36.6, and 36.6/40.0 at 90, 71, and 41 FPS, respectively.
Among the compared real-time detectors, RT-DETR-World-S+ reaches 36.6 Fixed AP, exceeding the best prior Fixed AP of 35.9 in this comparison by 0.7 points. RT-DETR-World-B further improves Fixed AP to 40.0, surpassing the same baseline by 4.1 points. It also obtains 37.1, 39.6, and 40.9 Fixed AP on rare, common, and frequent categories, respectively.
OV-DEIM-L remains a strong compact operating point, achieving 35.9 Fixed AP; RT-DETR-World-S+ and -B trade additional latency for higher accuracy.
Compared with general-purpose cross-modal models, RT-DETR-World-B remains 1.4 and 4.7 Fixed AP below MM-Grounding-DINO-T and LLMDet-T, respectively, delineating the gap under our streamlined inference architecture.

\setlength{\columnsep}{9pt}
\setlength{\intextsep}{4pt}
\begin{wraptable}{r}{0.37\textwidth}
\vspace{-5pt}
\centering
\tiny
\setlength{\tabcolsep}{1.0pt}
\renewcommand{\arraystretch}{0.96}
\caption{Zero-shot cross-domain transfer on ODinW.}
\label{tab:odinw}
\vspace{-9pt}
\begin{tabular}{@{}lccc@{}}
\toprule
Method & Backbone & ODinW13 & ODinW35 \\
\midrule
YOLO-Worldv2.1-S & YOLOv8-S  & 15.3 & 7.8 \\
YOLO-Worldv2.1-M & YOLOv8-M  & 21.4 & 10.5 \\
YOLO-Worldv2.1-L & YOLOv8-L  & 35.6 & 16.6 \\
\midrule
YOLOEv8-S & YOLOv8-S  & 28.0 & 12.7 \\
YOLOEv8-M & YOLOv8-M  & 30.9 & 14.3 \\
YOLOEv8-L & YOLOv8-L  & 31.9 & 14.8 \\
\midrule
YOLOEv11-S & YOLOv11-S & 28.2 & 12.7 \\
YOLOEv11-M & YOLOv11-M & 30.5 & 14.2 \\
YOLOEv11-L & YOLOv11-L & 31.4 & 14.4 \\
\midrule
OV-DEIM-S & ViT-T  & 32.9 & 14.8 \\
OV-DEIM-M & ViT-T+ & 36.6 & 16.2 \\
OV-DEIM-L & ViT-S  & 41.6 & 19.2 \\
\midrule
\rowcolor{blue!8}
RT-DETR-World-S
& DINOv3-S/16 & 40.1 & 18.7 \\
\rowcolor{blue!8}
RT-DETR-World-S+
& DINOv3-S+/16 & 41.5 & 18.8 \\
\rowcolor{blue!8}
RT-DETR-World-B
& DINOv3-B/16 & \textbf{43.8} & \textbf{20.2} \\
\bottomrule
\end{tabular}\vspace{-5pt}
\end{wraptable}

\paragraph{Cross-domain transfer on ODinW.}
ODinW comprises 35 detection datasets spanning diverse domains and vocabularies. Following prior work, we report mean AP over the commonly used 13-dataset subset (ODinW13) and the complete ODinW35 benchmark. As shown in Table~\ref{tab:odinw}, RT-DETR-World performs consistently well across all three model scales. RT-DETR-World-S achieves 40.1 and 18.7 AP, outperforming YOLO-Worldv2.1-L by 4.5 and 2.1 points and the strongest YOLOE variant by 8.2 and 3.9 points on ODinW13 and ODinW35, respectively. Despite using the smallest backbone in our family, it approaches OV-DEIM-L, which obtains 41.6 and 19.2 AP.
Scaling to RT-DETR-World-B raises performance to 43.8 AP on ODinW13 and 20.2 AP on ODinW35, surpassing OV-DEIM-L by 2.2 and 1.0 points, respectively. These results demonstrate that RT-DETR-World transfers effectively across datasets with substantially different domains, visual appearances, and category vocabularies.

\setlength{\columnsep}{9pt}
\setlength{\intextsep}{4pt}
\begin{wraptable}{r}{0.48\textwidth}
\vspace{-5pt}
\centering
\tiny
\setlength{\tabcolsep}{1.0pt}
\renewcommand{\arraystretch}{0.96}
\caption{COCO-O distribution-shift robustness. $^{\ddagger}$ indicates the use of COCO training images.}
\label{tab:coco-o}
\vspace{-8pt}
\begin{tabular}{@{}lcccc@{}}
\toprule
Method & Backbone & COCO AP & COCO-O AP & ER$\uparrow$ \\
\midrule
YOLO-Worldv2.1-S & YOLOv8-S
& 38.2 & 11.9 & $-5.3$ \\
YOLO-Worldv2.1-M & YOLOv8-M
& 43.8 & 16.7 & $-3.0$ \\
YOLO-Worldv2.1-L & YOLOv8-L
& 46.0 & 27.3 & ${+}6.6$ \\
\midrule
YOLOEv8-S & YOLOv8-S
& 35.6 & 15.4 & $-0.6$ \\
YOLOEv8-M & YOLOv8-M
& 42.2 & 17.7 & $-1.3$ \\
YOLOEv8-L & YOLOv8-L
& 45.4 & 19.8 & $-0.6$ \\
\midrule
YOLOEv11-S & YOLOv11-S
& 37.0 & 16.0 & $-0.7$ \\
YOLOEv11-M & YOLOv11-M
& 43.1 & 17.8 & $-1.6$ \\
YOLOEv11-L & YOLOv11-L
& 45.1 & 18.8 & $-1.5$ \\
\midrule
OV-DEIM-S & ViT-T
& 40.8 & 24.4 & ${+}6.0$ \\
OV-DEIM-M & ViT-T+
& 43.3 & 29.6 & ${+}10.2$ \\
OV-DEIM-L & ViT-S
& 45.9 & 43.3 & ${+}22.7$ \\
\midrule
\rowcolor{blue!8}
RT-DETR-World-S$^{\ddagger}$ & DINOv3-S/16
& 43.0 & 34.9 & ${+}15.6$ \\
\rowcolor{blue!8}
RT-DETR-World-S+$^{\ddagger}$ & DINOv3-S+/16
& 45.0 & 39.4 & ${+}19.2$ \\
\rowcolor{blue!8}
RT-DETR-World-B$^{\ddagger}$ & DINOv3-B/16
& 47.2 & 46.2 & \textbf{${+}24.9$} \\
\bottomrule
\end{tabular}\vspace{-5pt}
\end{wraptable}

\paragraph{Distribution-shift robustness on COCO-O.}
COCO-O retains the 80 COCO categories but evaluates them across six naturally shifted domains: sketch, weather, cartoon, painting, tattoo, and handmake. Because GroundingCapv2 includes COCO training images, we use COCO-O to evaluate distribution-shift robustness rather than unseen-category recognition. Following prior work, effective robustness is computed as
$\mathrm{ER}=
\mathrm{AP}_{\mathrm{COCO\text{-}O}}
-0.45\,\mathrm{AP}_{\mathrm{COCO}}$.

As shown in Table~\ref{tab:coco-o}, RT-DETR-World exhibits strong robustness across model scales. RT-DETR-World-S obtains 34.9 AP on COCO-O, exceeding YOLO-Worldv2.1-L by 7.6 points despite having a 3.0-point lower in-distribution COCO AP. It also surpasses the strongest YOLOE-L result by 15.1 points. RT-DETR-World-S+ further improves COCO-O AP to 39.4 and effective robustness to ${+}19.2$.
RT-DETR-World-B achieves the strongest overall results, with 47.2 AP on COCO, 46.2 AP on COCO-O, and ${+}24.9$ effective robustness. Compared with OV-DEIM-L, it improves COCO-O AP by 2.9 points and effective robustness by 2.2 points. These results show that RT-DETR-World retains strong detection performance under substantial appearance shifts rather than relying solely on in-distribution accuracy.

\subsection{Ablation Studies}
\label{sec:ablation}

Unless otherwise specified, all ablations use RT-DETR-World-S with a frozen DINOv3-S/16 backbone and identical training settings; AP denotes Fixed AP throughout.

\begin{table*}[t]
\centering
\caption{Ablation of training supervision on LVIS minival.}
\label{tab:component_ablation}
\vspace{-8pt}
\tiny
\setlength{\tabcolsep}{4.2pt}
\renewcommand{\arraystretch}{1.08}

\begin{tabular}{clccccc cccc}
\toprule
\multirow[c]{2}{*}{ID}
& \multirow[c]{2}{*}{Training Data}
& \multicolumn{2}{c}{LLM Teacher}
& \multicolumn{3}{c}{Deployable MiniLM Path}
& \multicolumn{4}{c}{LVIS Minival} \\
\cmidrule(lr){3-4}
\cmidrule(lr){5-7}
\cmidrule(lr){8-11}
&
&
Image
& Object
& Category
& Object Desc.
& Source C/P
& AP
& AP$_r$
& AP$_c$
& AP$_f$ \\
\midrule

B0
& GroundingCap-1M
& -- & -- & -- & -- & \checkmark
& 27.9 & 18.3 & 24.8 & 32.4 \\

B1
& GroundingCap-1M
& \checkmark & \checkmark & -- & -- & \checkmark
& 32.9 & 23.6 & 30.2 & 36.9 \\

\midrule

C0
& GroundingCapv2
& \checkmark & \checkmark & \checkmark & -- & --
& 33.1 & 25.7 & 32.2 & 35.1 \\

C1
& GroundingCapv2
& -- & -- & \checkmark & \checkmark & --
& 32.2 & 22.9 & 30.3 & 35.6 \\

C2
& GroundingCapv2
& \checkmark & -- & \checkmark & \checkmark & --
& 33.1 & 27.8 & 32.6 & 34.4 \\

C3
& GroundingCapv2
& -- & \checkmark & \checkmark & \checkmark & --
& 33.6 & 29.1 & 32.5 & 35.3 \\

\rowcolor{blue!8}
\textbf{C4}
& \textbf{GroundingCapv2}
& \checkmark & \checkmark & \checkmark & \checkmark & --
& \textbf{35.2}
& \textbf{32.0}
& \textbf{33.8}
& \textbf{36.9} \\

\bottomrule
\end{tabular}

\vspace{2pt}
\begin{minipage}{0.98\textwidth}
\tiny
\textbf{Notes:}
GroundingCap-1M denotes LLMDet's annotations, while GroundingCapv2 denotes our tri-level annotations.
Image and Object denote LLM-teacher alignment with image and object descriptions, respectively.
Category and Object Desc.\ denote MiniLM supervision from category names and object descriptions.
Source C/P denotes source-specific category or region-phrase supervision.
The LLM teacher and auxiliary alignment modules are removed at inference.
\end{minipage}
\vspace{-10pt}
\end{table*}

\paragraph{Supervision ablation.}
Table~\ref{tab:component_ablation} analyzes the contributions of the two DDA pathways under different annotation formulations. On GroundingCap-1M annotations, adding object- and image-level LLM-teacher supervision improves AP from 27.9 (B0) to 32.9 (B1), a gain of 5.0 points without changing the deployable detector. Under GroundingCapv2, adding MiniLM object-description alignment to the category-only configuration improves AP from 33.1 (C0) to 35.2 (C4), with a large AP$_r$ gain from 25.7 to 32.0. Starting from the MiniLM-only configuration C1, image- and object-level teacher alignment individually improve AP from 32.2 to 33.1 (C2) and 33.6 (C3), respectively. Combining both levels raises AP to 35.2 (C4). The complementary gains from image- and object-level targets demonstrate the benefit of transferring both global context and instance semantics, while the strong rare-category improvement supports the use of multi-level semantic supervision.

\paragraph{Comparison of auxiliary alignment strategies.}

\setlength{\columnsep}{10pt}
\setlength{\intextsep}{4pt}
\begin{wraptable}{r}{0.39\textwidth}
\vspace{-4pt}
\centering
\tiny
\caption{Auxiliary alignment strategies on LVIS minival.}
\label{tab:auxiliary_method_comparison}
\vspace{-8pt}
\setlength{\tabcolsep}{2.2pt}
\renewcommand{\arraystretch}{1.04}
\begin{tabular}{lcccc}
\toprule
Auxiliary Supervision
& AP
& AP$_r$
& AP$_c$
& AP$_f$ \\
\midrule

LLMDet-style Caption Generation
& 28.5 & 18.3 & 26.4 & 32.2 \\

Hard Contrastive Alignment
& 33.6 & 27.6 & 31.7 & 36.4 \\

SoftCLIP-style Soft Alignment
& 33.8 & 25.1 & 32.6 & 36.5 \\

SRCL-style Similarity Reweighting
& 33.9 & 26.2 & 32.2 & 36.8 \\

\rowcolor{blue!8}
\textbf{RNR (Ours)}
& \textbf{35.2}
& \textbf{32.0}
& \textbf{33.8}
& \textbf{36.9} \\

\bottomrule
\end{tabular}

\vspace{2pt}
\begin{minipage}{0.96\linewidth}
\tiny
\textit{Note:} SoftCLIP and SRCL are adapted to the same auxiliary-alignment setting.
\end{minipage}
\vspace{-3pt}
\end{wraptable}

Table~\ref{tab:auxiliary_method_comparison} compares RNR with alternative strategies for auxiliary semantic supervision. The LLMDet-style caption-generation objective~\citep{fu2025llmdet} performs poorly under the same compact detector setting, whereas hard contrastive alignment provides a stronger baseline at 33.6 AP. SoftCLIP-style soft alignment~\citep{gao2024softclip} and SRCL-style similarity regulation~\citep{jiang2023vision} yield modest improvements, reaching 33.8 and 33.9 AP, respectively.
Unlike SoftCLIP, which replaces exact correspondence targets with soft targets derived from intra-modal similarity, and SRCL, which relies on progressively refined cross-modal similarity, RNR preserves exact-positive targets and only discounts related negatives using frozen LLM-teacher relations. This design separates semantic relatedness from instance correspondence, avoiding ambiguous positive targets while providing stable relation estimates. Consequently, RNR achieves 35.2 AP, outperforming hard contrastive alignment and SRCL-style regulation by 1.6 and 1.3 points, respectively. The improvement is most pronounced for rare categories, where RNR raises AP$_r$ by 4.4 points over hard alignment.
\textit{\textbf{See Appendices~\ref{sec:Hyperparameter} and~\ref{sec:Visualization} for hyperparameter sensitivity and qualitative results, respectively.}}
\section{Conclusion}
We present \textbf{RT-DETR-World}, a compact open-vocabulary detector that learns the rich semantics conveyed by descriptions during training while retaining lightweight query--text matching at inference. \textbf{GroundingCapv2} provides category-, object-, and image-level supervision. \textbf{DDA} transfers fine-grained instance semantics and scene context through a deployable MiniLM pathway and offline LLM-teacher targets, while \textbf{RNR} relaxes semantically related negatives. All teacher-side components are removed after training, adding no inference overhead. Extensive experiments demonstrate competitive large-vocabulary detection, strong cross-domain transfer, and distribution-shift robustness with real-time inference. These results indicate that rich training-time language supervision can enhance compact open-vocabulary detectors without increasing inference cost.

\subsection*{AI use statement}

In this work, we used generative AI tools for dataset construction and translation. Specifically, Qwen3-VL was used in our GroundingAgent-Opt pipeline to generate and curate annotations for GroundingCapv2, including instance descriptions and category annotations. The models, agent roles, routing strategy, and data-curation procedure are described in detail in the paper. Generative AI tools were also used to assist with English translation of author-written content.

We did not use generative AI tools to develop the proposed methodology, formulate theoretical or mathematical claims, write proofs, design experiments, implement the method, or interpret experimental results. The remaining required disclosure tasks are not applicable to this work.

Additionally, generative AI tools were used for language polishing and readability improvement of author-written text. All AI-assisted writing was reviewed and revised by the authors, and AI-generated annotations were processed and verified through the curation and quality-control procedures described in the paper. We take responsibility for the final content of this work, including text, claims, and artifacts produced with the aid of generative AI.



\subsection*{Ethics statement}
This work does not involve the collection of new human-subject data or personally identifiable information. All experiments are conducted on publicly available object detection, grounding, and image--text datasets or their processed derivatives. GroundingCapv2 is constructed by reorganizing and augmenting existing data with model-generated annotations, followed by the curation and quality-control procedures described in the paper. We do not intend the proposed method or dataset construction pipeline for applications that violate privacy, safety, or applicable data-use policies.

\subsection*{Reproducibility statement}
We provide detailed descriptions of the model architecture, training settings, evaluation protocols, data construction procedure, and ablation configurations in the paper. The construction and standardization of GroundingCapv2 are also described in the main text. To facilitate reproducibility, we will make the code and processed dataset annotations publicly available.

\bibliography{iclr2027_conference}
\bibliographystyle{iclr2027_conference}

\newpage

\appendix

\section*{Appendix}
\section{Related work} 
\label{sec:related_work} 

\paragraph{Zero-shot detection (ZSD) and open-vocabulary detection (OVD).}
Early ZSD~\citep{bansal2018zero} transfers semantic knowledge from seen to predefined unseen categories through visual--semantic embeddings, while OVR-CNN~\citep{zareian2021open} extends this paradigm to OVD via image--caption pretraining. With large-scale VLMs such as CLIP~\citep{radford2021learning}, later methods mainly explore knowledge transfer and weak supervision. ViLD~\citep{gu2021open}, OADP~\citep{wang2023object}, and DK-DETR~\citep{li2023distilling} distill region-level or relational knowledge; F-VLM~\citep{kuo2022f} and NoOVD~\citep{zhang2026noovd} adapt pretrained VLM representations to detection; while Detic~\citep{zhou2022detecting} and SAS-Det~\citep{zhao2024taming} expand the training vocabulary through image-level supervision or pseudo-labels.
Grounded pretraining provides another effective paradigm. GLIP~\citep{li2022grounded}, Grounding DINO~\citep{liu2024grounding}, and MM-Grounding-DINO~\citep{zhao2024open} unify detection and phrase grounding, while LLMDet~\citep{fu2025llmdet} further introduces image- and region-level caption supervision. Efficient OVD is explored by Grounding DINO 1.5-Edge~\citep{ren2024grounding} for edge deployment, YOLO-World~\citep{cheng2024yolo} and YOLOE~\citep{wang2025yoloe} via efficient or re-parameterizable vision--language alignment, YOLO-UniOW~\citep{liu2024yolo} via adaptive text decision learning, and OV-DEIM~\citep{wang2026ov} with an efficient end-to-end DETR design. 
In contrast, RT-DETR-World uses object- and image-level descriptions only during training, preserves semantic relations through RNR, and removes all teacher-side components at inference.

\paragraph{Multi-granular description supervision and relational alignment.}
Conventional image--text contrastive learning treats all unmatched pairs as negatives, ignoring their many-to-many semantic relations. CLIP-PSD~\citep{andonian2022robust} and SoftCLIP~\citep{gao2024softclip} introduce soft targets via self-distillation or intra-modal similarities, CyCLIP~\citep{goel2022cyclip} regularizes cross-modal geometry, and SRCL~\citep{jiang2023vision} reweights negatives by semantic similarity. These methods mainly target general vision--language pretraining and derive relational supervision from model predictions or external visual--language signals rather than sample-specific descriptions.

In OVD, DK-DETR~\citep{li2023distilling} distills region semantics and relations, RALF~\citep{kim2024retrieval} exploits retrieved negatives and LLM-generated concepts, and HD-OVD~\citep{fu2025hierarchical} transfers CLIP knowledge across instance, category, and image levels. BARON~\citep{wu2023aligning} aligns related region groups, while DeCo-DETR~\citep{li2026deco} organizes LLaVA-generated region descriptions into hierarchical prototypes. In contrast, we use sample-specific object and image descriptions encoded by a frozen LLM2CLIP~\citep{huang2026llm2clip} teacher to supervise matched queries and global visual representations. Their semantic similarities relax related negatives while preserving exact positives. All teacher features are precomputed, leaving inference unchanged.


\section{Hyperparameter Sensitivity}
\label{sec:Hyperparameter}

We vary the LLM-teacher weight, RNR relaxation coefficient, and MiniLM description weight individually while fixing the remaining hyperparameters to their default values. All experiments use RT-DETR-World-S and report Fixed AP on LVIS minival.

\begin{table*}[h]
\centering
\caption{Hyperparameter sensitivity on LVIS minival. All results are reported using Fixed AP.}
\label{tab:hyperparameter_sensitivity}
\vspace{-7pt}

\captionsetup[subtable]{
    font=footnotesize,
    labelfont=bf,
    justification=centering,
    singlelinecheck=true,
    skip=3pt
}

\begingroup
\scriptsize
\setlength{\tabcolsep}{2.8pt}

\begin{subtable}[t]{0.31\textwidth}
\centering
\caption{LLM teacher weight $\lambda_{\mathrm{aux}}$}
\label{tab:sensitivity_aux_weight}
\vspace{-3pt}
\begin{tabular}{ccccc}
\toprule
$\lambda_{\mathrm{aux}}$
& AP & $\mathrm{AP}_r$ & $\mathrm{AP}_c$ & $\mathrm{AP}_f$ \\
\midrule
0
& 32.2 & 22.9 & 30.3 & 35.6 \\
0.05
& 31.6 & 23.8 & 29.3 & 35.0 \\
\rowcolor{blue!8}
\textbf{0.1}
& \textbf{35.2}
& 32.0
& \textbf{33.8}
& \textbf{36.9} \\
0.2
& 34.3
& \textbf{32.5}
& 33.0
& 35.8 \\
\bottomrule
\end{tabular}
\end{subtable}
\hfill
\begin{subtable}[t]{0.32\textwidth}
\centering
\caption{RNR coefficient $\alpha$}
\label{tab:sensitivity_relaxation}
\vspace{-3pt}
\begin{tabular}{ccccc}
\toprule
$\alpha$
& AP & $\mathrm{AP}_r$ & $\mathrm{AP}_c$ & $\mathrm{AP}_f$ \\
\midrule
0
& 33.6 & 27.6 & 31.7 & 36.4 \\
0.1
& 34.3 & 30.6 & 33.4 & 35.4 \\
\rowcolor{blue!8}
\textbf{0.25}
& \textbf{35.2}
& \textbf{32.0}
& \textbf{33.8}
& \textbf{36.9} \\
0.5
& 33.4 & 29.4 & 32.0 & 35.4 \\
\bottomrule
\end{tabular}
\end{subtable}
\hfill
\begin{subtable}[t]{0.35\textwidth}
\centering
\caption{MiniLM description weight $\lambda_{\mathrm{desc}}$}
\label{tab:sensitivity_desc_weight}
\vspace{-3pt}
\begin{tabular}{ccccc}
\toprule
$\lambda_{\mathrm{desc}}$
& AP & $\mathrm{AP}_r$ & $\mathrm{AP}_c$ & $\mathrm{AP}_f$ \\
\midrule
0
& 33.1 & 25.7 & 32.2 & 35.1 \\
0.1
& 33.9 & 28.0 & 33.1 & 35.6 \\
\rowcolor{blue!8}
\textbf{0.25}
& \textbf{35.2}
& \textbf{32.0}
& 33.8
& \textbf{36.9} \\
0.5
& 34.3 & 30.2 & \textbf{34.1} & 35.1 \\
0.75
& 33.3 & 29.2 & 32.3 & 34.8 \\
1.0
& 33.3 & 29.7 & 33.3 & 34.0 \\
\bottomrule
\end{tabular}
\end{subtable}

\endgroup

\vspace{2pt}
\begin{minipage}{0.98\textwidth}
\tiny
\textit{Notes:}
$\lambda_{\mathrm{aux}}=\lambda_{\mathrm{o}}=\lambda_{\mathrm{g}}$
and $\lambda_{\mathrm{desc}}=\lambda_{\mathrm{M}}$.
Unless varied, the default values are
$\lambda_{\mathrm{aux}}=0.1$, $\alpha=0.25$, and
$\lambda_{\mathrm{desc}}=0.25$.
\end{minipage}
\end{table*}

As shown in Table~\ref{tab:hyperparameter_sensitivity}, $\lambda_{\mathrm{aux}}=0.1$ achieves the highest overall AP of 35.2. Increasing it to 0.2 further improves $\mathrm{AP}_r$ to 32.5 but reduces overall AP to 34.3.

For RNR, increasing $\alpha$ from 0 to 0.25 improves AP from 33.6 to 35.2 and $\mathrm{AP}_r$ from 27.6 to 32.0, whereas excessive relaxation at $\alpha=0.5$ degrades performance. Similarly, increasing the MiniLM description weight from 0 to 0.25 improves AP from 33.1 to 35.2 and $\mathrm{AP}_r$ from 25.7 to 32.0. Although $\lambda_{\mathrm{desc}}=0.5$ yields the highest $\mathrm{AP}_c$ of 34.1, larger weights reduce overall AP to 33.3 at both 0.75 and 1.0. We therefore use $\lambda_{\mathrm{aux}}=0.1$, $\alpha=0.25$, and $\lambda_{\mathrm{desc}}=0.25$ by default.


\section{Visualization and Analysis}
\label{sec:Visualization}

\begin{figure*}[t]
	\centering
	\includegraphics[width=1.0\linewidth]{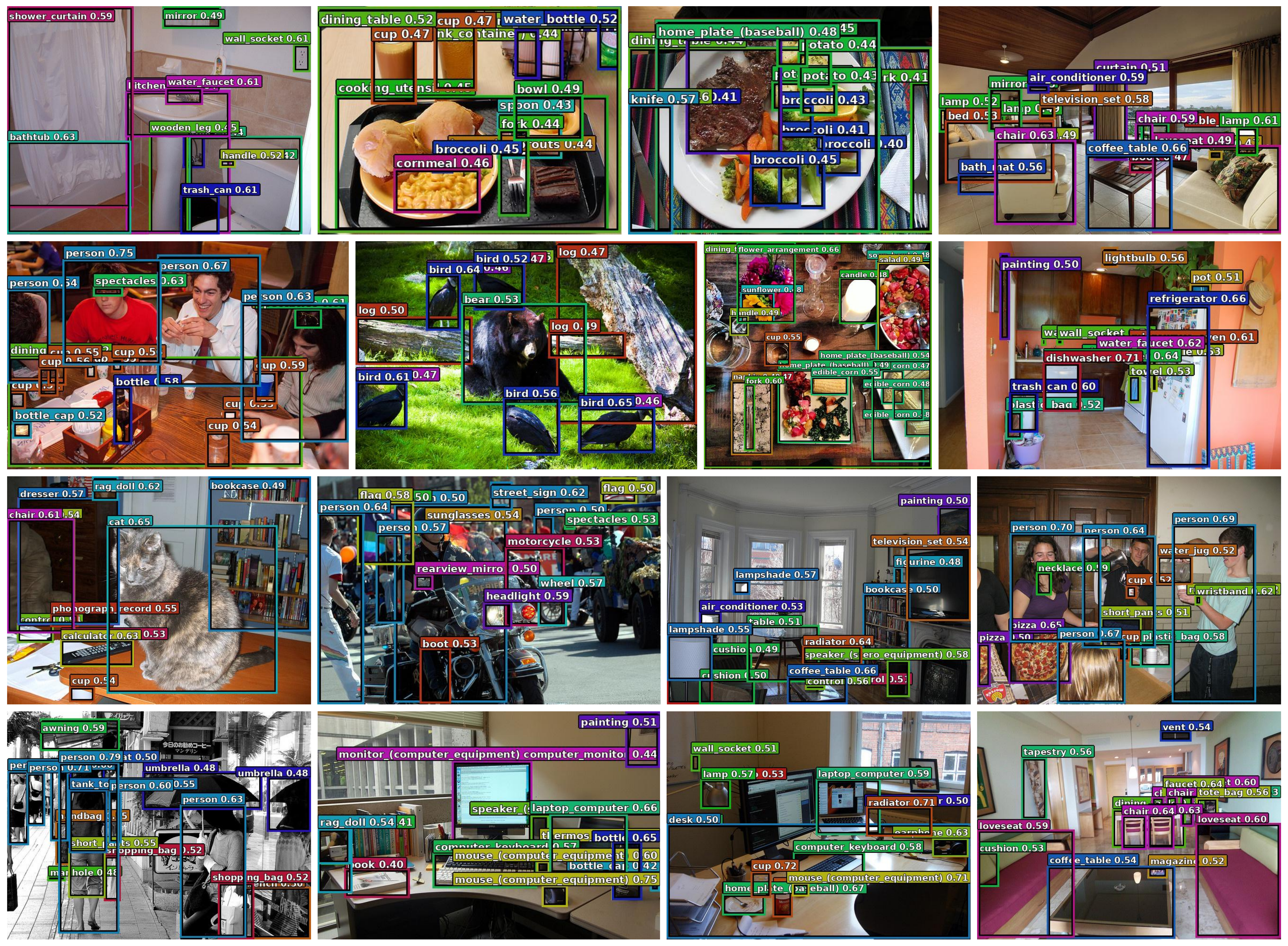}\vspace{-8pt}
	\caption{\textbf{Qualitative zero-shot detections on LVIS minival.} Using the full 1,203-category vocabulary, RT-DETR-World localizes diverse objects across indoor, outdoor, and cluttered scenes, including small and fine-grained instances. Labels and numbers indicate predicted categories and confidence scores, respectively.}
    \label{fig:qualitative}  %
	\vspace{-0pt}
\end{figure*}

\paragraph{Qualitative results.}
Fig.~\ref{fig:qualitative} presents zero-shot detections produced by RT-DETR-World-B using all 1,203 LVIS category names as queries. The model recognizes objects across diverse environments, ranging from indoor living spaces and dining scenes to crowded streets and natural settings. It successfully localizes not only salient objects such as persons, furniture, and animals, but also small or fine-grained instances such as wall sockets, water faucets, bottle caps, wristbands, radiators, and computer peripherals. The results also show consistent localization of multiple instances in crowded scenes, as illustrated by the birds, persons, cups, and tabletop objects. These examples qualitatively support the model's ability to associate category concepts with visual cues at different scales and levels of granularity. Some low-confidence and overlapping predictions remain in densely populated scenes, indicating that fine-grained category discrimination and confidence calibration remain challenging under a large vocabulary.

\end{document}

%% file: math_commands.tex
\usepackage{amsmath,amsfonts,bm}

\def\eqref#1{equation~\ref{#1}}

\def\1{\bm{1}}

\DeclareMathAlphabet{\mathsfit}{\encodingdefault}{\sfdefault}{m}{sl}
\SetMathAlphabet{\mathsfit}{bold}{\encodingdefault}{\sfdefault}{bx}{n}

